\documentclass[10pt,twocolumn,letterpaper]{article}

\usepackage{wacv}              

\usepackage{graphicx}
\usepackage{amsmath}
\usepackage{amssymb}
\usepackage{booktabs}

\usepackage[accsupp]{axessibility} 

\usepackage[pagebackref,breaklinks,colorlinks]{hyperref}

\usepackage{graphicx}
\usepackage{amsmath}
\usepackage{amssymb}
\usepackage{gensymb}

\usepackage{amssymb}
\usepackage{xcolor}

\usepackage{subcaption}

\usepackage{kotex}
\usepackage{color}
\usepackage{array}

\usepackage{bm}
\usepackage{nth}

\newcolumntype{C}[1]{>{\centering\let\newline\\\arraybackslash\hspace{0pt}}m{#1}}
\newcolumntype{L}[1]{>{\let\newline\\\arraybackslash\hspace{0pt}}m{#1}}

\usepackage{rotating} 

\usepackage{amsmath}

\usepackage{booktabs}
\usepackage{listings}
\usepackage{algorithm}
\usepackage{algorithmic}
\usepackage{mathtools, nccmath}

\usepackage{enumitem}

\usepackage[capitalize]{cleveref}
\crefname{section}{Sec.}{Secs.}
\Crefname{section}{Section}{Sections}
\Crefname{table}{Table}{Tables}
\crefname{table}{Tab.}{Tabs.}

\usepackage[capitalize]{cleveref}
\crefname{section}{Sec.}{Secs.}
\Crefname{section}{Section}{Sections}
\Crefname{table}{Table}{Tables}
\crefname{table}{Tab.}{Tabs.}

\def\wacvPaperID{1262} 
\def\confName{WACV}
\def\confYear{2025}

\begin{document}

\title{Beta Sampling is All You Need: Efficient Image Generation Strategy for Diffusion Models using Stepwise Spectral Analysis}

\author{
Haeil Lee\thanks{These authors contributed equally to this work.} \and
Hansang Lee\footnotemark[1] \and
Seoyeon Gye \and
Junmo Kim \and\\
School of Electrical Engineering, KAIST, South Korea\\
{\tt\small \{haeil.lee, hansanglee, sawyun, junmo.kim\}@kaist.ac.kr}
}
\maketitle

\begin{abstract}
Generative diffusion models have emerged as a powerful tool for high-quality image synthesis, yet their iterative nature demands significant computational resources. This paper proposes an efficient time step sampling method based on an image spectral analysis of the diffusion process, aimed at optimizing the denoising process. Instead of the traditional uniform distribution-based time step sampling, we introduce a Beta distribution-like sampling technique that prioritizes critical steps in the early and late stages of the process. Our hypothesis is that certain steps exhibit significant changes in image content, while others contribute minimally. We validated our approach using Fourier transforms to measure frequency response changes at each step, revealing substantial low-frequency changes early on and high-frequency adjustments later. Experiments with ADM and Stable Diffusion demonstrated that our Beta Sampling method consistently outperforms uniform sampling, achieving better FID and IS scores, and offers competitive efficiency relative to state-of-the-art methods like AutoDiffusion. This work provides a practical framework for enhancing diffusion model efficiency by focusing computational resources on the most impactful steps, with potential for further optimization and broader application.
\end{abstract}

\section{Introduction}
\label{sec:intro}

\begin{figure}[t]
  \centering
  \begin{subfigure}{0.90\linewidth}
    \includegraphics[width=\linewidth]{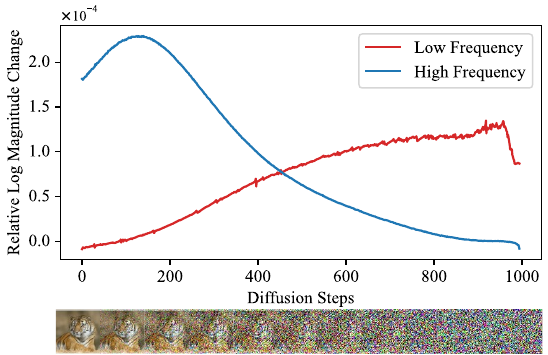}
    \caption{Spectral analysis of the denoising process}
    \label{fig:teaser-a}
  \end{subfigure}
  \begin{subfigure}{0.90\linewidth}
    \includegraphics[width=\linewidth]{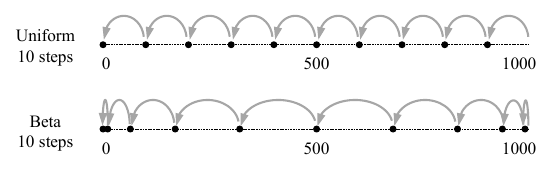}
    \caption{Uniform and Beta Sampling for denoising process}
    \label{fig:teaser-b}
  \end{subfigure}
    \begin{subfigure}{0.90\linewidth}
    \includegraphics[width=\linewidth]{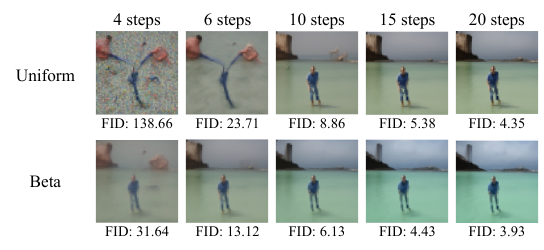}
    \caption{Samples generated by ADM-G with uniform and Beta Sampling}
    \label{fig:teaser-c}
  \end{subfigure}
  \caption{\textbf{An Overview.} 
  (a) We analyzed the Fourier transform of images generated at each time step during the denoising process and found that the changes in low- and high-frequency components are concentrated in the early and the later stages, respectively. 
  (b) Based on this, we propose a Beta distribution-like sampling method that focuses on key stages with significant frequency changes.
  (c) Experiments show our method generates higher quality images at lower steps compared to uniform sampling.
  }
  \label{fig:teaser}
\end{figure}

Generative diffusion models have emerged as a powerful tool for high-quality image generation, producing results that rival or surpass traditional generative adversarial networks (GANs)~\cite{dhariwal2021diffusion_adm}. 
These models iteratively refine images through a diffusion process, where noise is progressively added and then removed, ultimately generating realistic images from random noise~\cite{ho2020ddpm}. 
However, this iterative nature comes with a significant computational cost, necessitating numerous time steps to achieve high-quality outputs. 
Efficient methods are therefore crucial to reduce computational burden while maintaining the quality of generated images.


Previous efforts to improve the efficiency of diffusion models have focused on reducing the number of sampling steps in denoising process to improve the efficiency of diffusion models.
Some approaches model the sampling process as ordinary differential equations (ODEs), which enables fewer steps using first-order and higher-order solvers\cite{song2021ddim, liu2022pndm_plms, lu2022dpmsolver}.
Another effective strategy is to use knowledge distillation techniques repeatedly to condense multiple steps into a single step \cite{salimans2022progressive, berthelot2023tract, meng2023distillation}. This enables the generation of high-quality images in 10 steps or fewer.
Recently, new approaches have been proposed based on the idea that the sampling process of diffusion models plays a different role at each step\cite{rombach2022ldm, balaji2023ediffi, yang2022spectral_diffusion}. Some works aim to reduce the number of sampling steps by selecting only the optimal step for particular diffusion models\cite{watson2022ddss, li2023autodiffusion, fang2023diffpruning}, while others aim to increase efficiency by using multiple smaller but step-specific models for each step\cite{yang2022spectral_diffusion, li2023autodiffusion, yang2024ddsm, lee2023meme}.

In this paper, we propose a novel Beta Sampling method for improving the efficiency of generative diffusion models. 
By conducting an image spectral analysis of the diffusion process using Fourier transform, we identified that significant changes in image content occur predominantly at the early and late stages of the denoising process.
Based on this insight, we introduce a Beta distribution-like time step sampling method that emphasizes critical steps in the early and late stages of the process.
Our key hypothesis is that certain steps in the diffusion process exhibit significant changes in image content, while others contribute minimally. 
By focusing on these impactful time steps, we aim to enhance the efficiency of image generation without compromising quality, in contrast to traditional uniform sampling.
We validate our approach through experiments with ADM and Stable Diffusion models, demonstrating that our Beta Sampling method consistently achieves better FID and IS scores compared to uniform sampling and exhibits competitive efficiency against the state-of-the-art AutoDiffusion method. 
Our findings highlight the potential for substantial computational savings and quality improvements in image generation tasks.

Our contributions are as follows.
\begin{itemize}[itemsep=0em]
    \item We provide a spectral analysis of the diffusion process through Fourier transform, identifying that the significant changes in frequency component are concentrated in the early and later stages of the denoising process.
    \item Based on our analysis, we introduce a Beta distribution-like sampling method that prioritizes steps with substantial changes in low and high-frequency components.
    \item We demonstrate through experiments with ADM and Stable Diffusion that our proposed sampling method consistently achieves improved FID and IS scores compared to uniform sampling, and offers competitive efficiency against state-of-the-art methods like AutoDiffusion in terms of computational complexity.
\end{itemize}

\section{Related works}
\label{sec:relwork}

\subsection{Denoising process of diffusion models}

There is an active research effort to analyze and improve the denoising process of diffusion models from various aspects. So far, most analyses agree that the diffusion model's denoising process proceeds in a coarse-to-fine manner, i.e., it generates the overall structure of the image by focusing on low-frequency components at the beginning, and then completes the details through changes in high-frequency components at the end, resulting in progressively better images.

Several efforts have been focused on analyzing the denoising process of the diffusion models. First, Choi \etal\cite{choi2022perception} analyzed the process of adding noise to an image using LPIPS distance, with the idea that the diffusion model has a different pretask for each step, and proposed that the denoising process of the diffusion model can be divided into three stages: 1) creating coarse features 2) generating perceptually rich contents 3) and removing remaining noise.

To analyze how the latent structure of the diffusion model varies with the diffusion timestep, Park \etal\cite{park2023understanding} identified the frequency domain of the local latent basis by power spectral density (PSD) analysis, and confirmed the shift from low-frequency to high-frequency as the denoising process progresses.
By analyzing the frequency domain of the image obtained at each step, we also confirmed that the low-frequency part is restored at the beginning and the high-frequency detail is restored at the end\cite{lee2023meme,  yang2022spectral_diffusion}.
Furthermore, Li \etal\cite{li2023autodiffusion} argued that the difficulty and importance of each of these different steps is different, and that efficient sampling can be achieved by finding an optimal time step.

More recently, researchers have tried to visually interpret the denoising process through a text to image diffusion model\cite{park2024explaining}. This paper investigated the spatial recovery level at each timestep in the denoising process and showed that the focal region of the model changes from semantic information to fine-grained regions. 

\subsection{Frequency Analysis on Diffusion Models}

Frequency analysis has long been a fundamental tool for image processing and analysis in the field of computer vision.
It enables the separation of high and low-frequency components of an image.
The Fourier transform has been widely used for frequency analysis in identifying major patterns in images, removing noise, and compressing images.

Research has shown that deep neural networks initially adapt to low-frequency signals, while learning high-frequency details more gradually\cite{xu2019frequency, xu2019training, rahaman2018spectral}. This phenomenon, known as \textit{spectral bias}, has also been observed in deep generative models such as GANs\cite{schwarz2021frequency, khayatkhoei2022spatial, frank2020leveraging}.
A similar spatial frequency bias has been found in diffusion models. 
Choi~\etal\cite{choi2022perception} suggests that the denoising process of diffusion models shows three distinguishable phases. The first phase captures coarse attributes, while the subsequent phases progressively incorporate finer details. Other studies have noted that throughout the progression of the denoising process, low-frequency components are preserved, while high-frequency components undergo rapid changes\cite{si2023freeu} or the ratio of high-frequency increases as the denoising progresses\cite{park2023understanding}.

This observation indicates that the denoising task of diffusion models varies at each time step, leading to distinct characteristics in spectral analysis at each step.
Recent studies have utilized this fact to enhance the quality of generated samples \cite{si2023freeu} or computational efficiency \cite{yang2022spectral_diffusion, lee2023meme}. For example, some studies give more weight to a specific frequency at each step\cite{yang2022spectral_diffusion, si2023freeu}, while Lee~\etal\cite{lee2023meme} utilizes different smaller models at each step, each specialized for a particular range of frequencies.

\section{Spectral Analysis of Denoising Processes in Diffusion Models}
\label{sec:analysis}

The denoising process in diffusion models has been demonstrated to encompass several distinct stages, each characterized by unique model behaviors~\cite{deja2022analyzing,choi2022perception}. This variability in behavior across time steps suggests that the importance of each step in the denoising process may not be uniform. 
To investigate the relative importance of each step in the denoising process, we conducted a comprehensive spectral analysis of the denoising procedure in pre-trained diffusion models. This analytical approach allows us to inspect which parts of the denoising process contribute most significantly to meaningful changes in the generated image. By decomposing the process into its spectral components, we aim to provide insights into the differential contributions of each denoising step to the final output quality.

\noindent
\textbf{Diffusion Models.}
To investigate the frequency characteristics of diffusion models, we conducted a comprehensive analysis on two prominent models: ADM-G~\cite{dhariwal2021diffusion_adm} and Stable Diffusion~\cite{rombach2022ldm}. The ADM-G model was trained on ImageNet 64$\times$64~\cite{imagenet}, while Stable Diffusion was trained on the LAION-5B~\cite{schuhmann2022laion} dataset. 

For the ADM-G model, we observed the denoising process for a total of 10,000 image generations. Specifically, we generated 10 images for each class in the ImageNet dataset, ensuring a broad representation across all categories. 
In the case of Stable Diffusion, we randomly selected 1,000 captions from the validation set of COCO~\cite{lin2014microsoft_coco} dataset to serve as prompts for image generation.

\begin{figure}[t]
  \centering
  \begin{subfigure}{0.45\linewidth}
    \includegraphics[width=\linewidth]{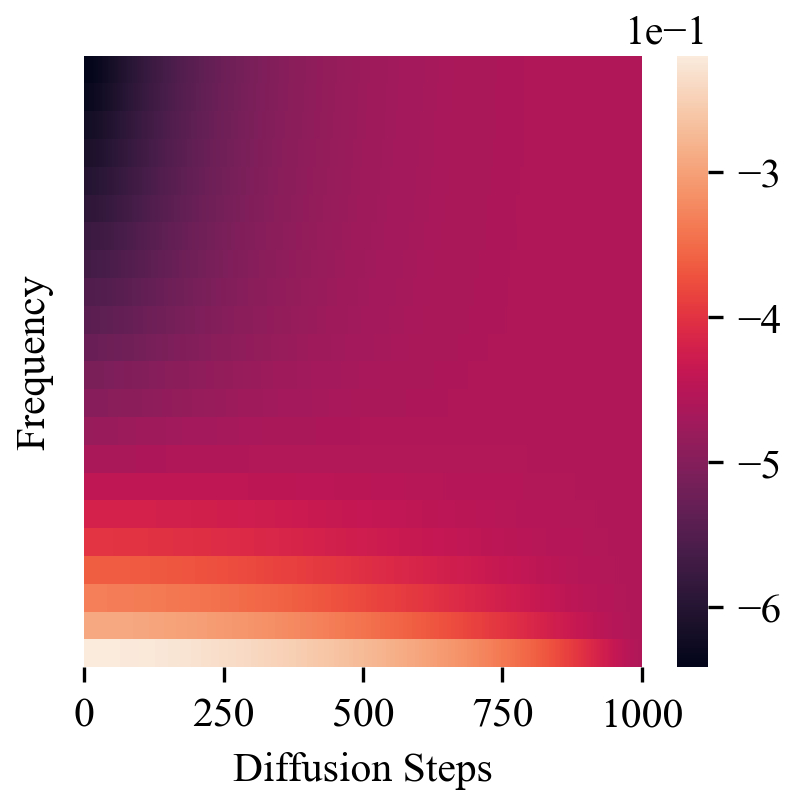}
    \caption{Relative frequency in ADM-G}
    \label{fig:freq_adm}
  \end{subfigure}
  \begin{subfigure}{0.45\linewidth}
    \includegraphics[width=\linewidth]{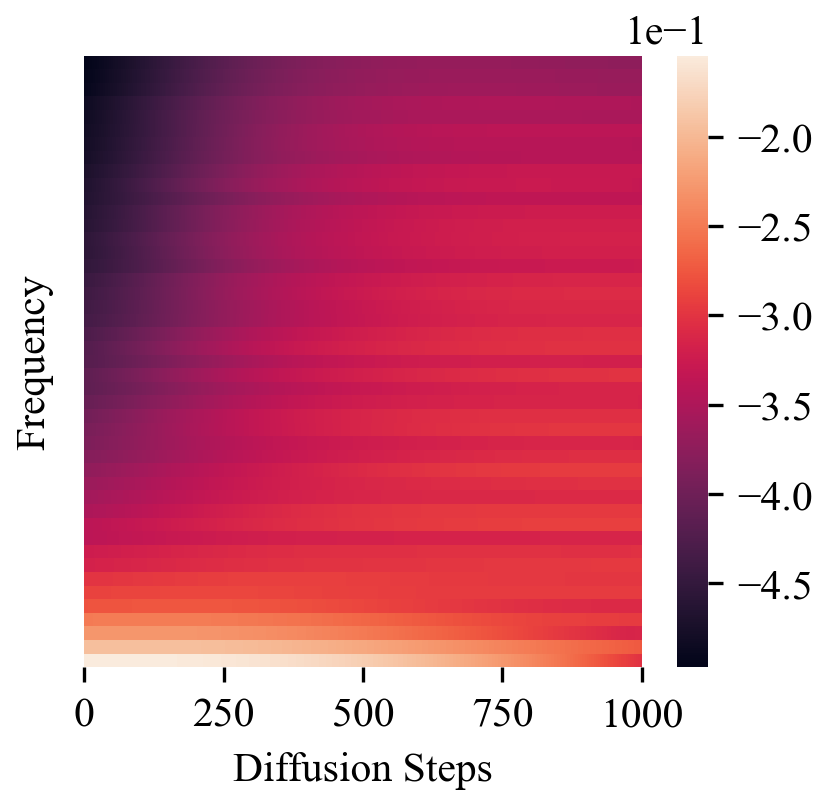}
    \caption{Relative frequency in SD}
    \label{fig:freq_SD}
  \end{subfigure}
\begin{subfigure}{0.45\linewidth}
    \includegraphics[width=\linewidth]{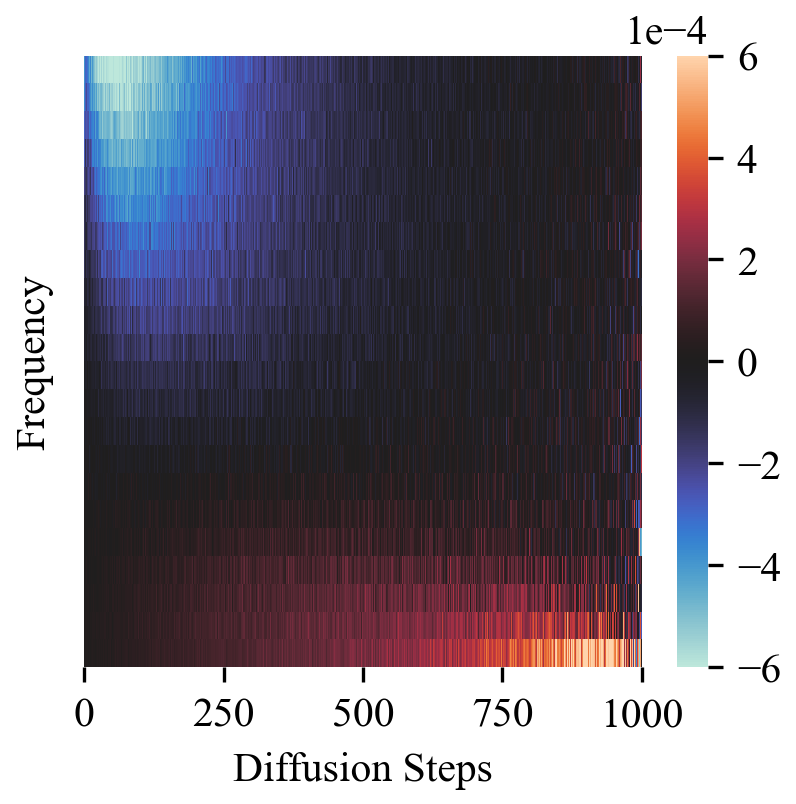}
    \caption{Frequency change in ADM-G}
    \label{fig:freq_change_ADM}
  \end{subfigure}
  \begin{subfigure}{0.45\linewidth}
    \includegraphics[width=\linewidth]{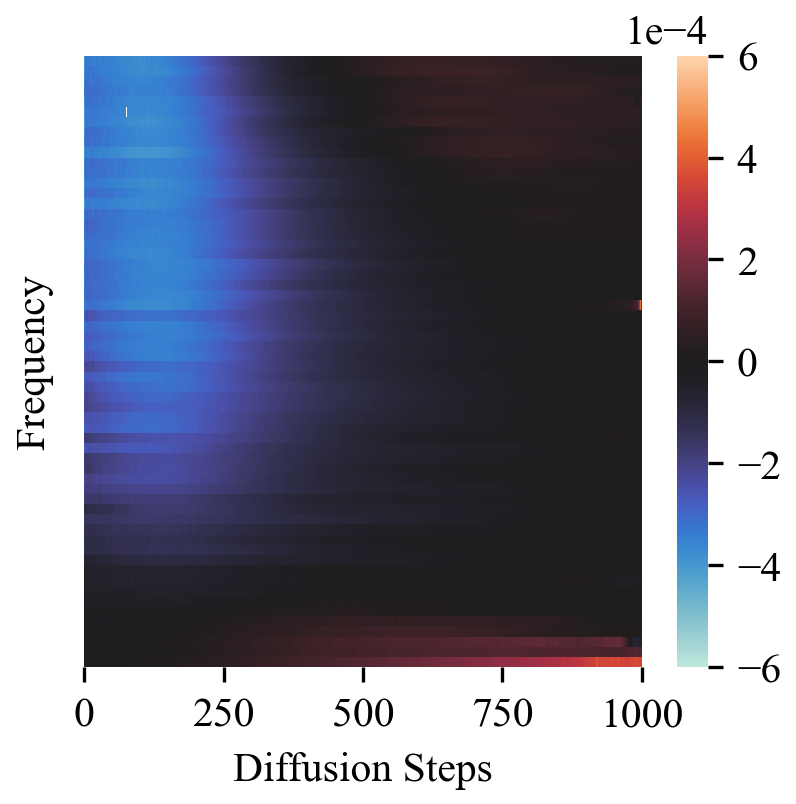}
    \caption{Frequency change in SD}
    \label{fig:freq_change_SD}
  \end{subfigure}
  \begin{subfigure}{0.45\linewidth}
    \includegraphics[width=\linewidth]{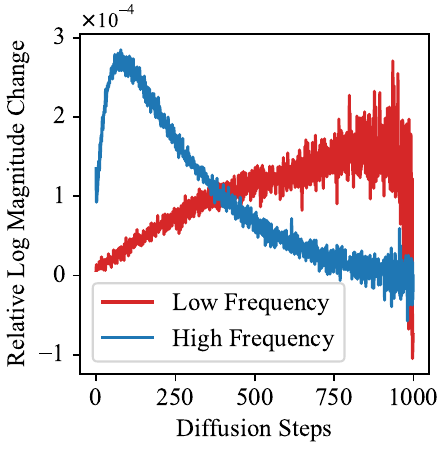}
    \caption{Average of change in ADM-G}
    \label{fig:avg_change_adm}
  \end{subfigure}  
  \begin{subfigure}{0.45\linewidth}
    \includegraphics[width=\linewidth]{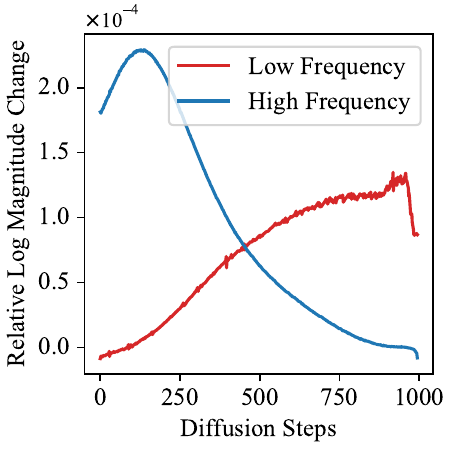}
    \caption{Average of change in SD}
    \label{fig:avg_change_SD}
  \end{subfigure}
  \caption{Spectral analysis of denoising process in ADM-G~\cite{dhariwal2021diffusion_adm} and Stable Diffusion (SD)~\cite{rombach2022ldm}. The trend in the changes of high-frequency and low-frequency components during the denoising steps demonstrates that the core of the diffusion model’s image denoising process lies in the early and late stages.}
  \label{fig:frequency_change}
\end{figure}

\noindent
\textbf{Spectral Analysis.}
To conduct spectral analysis of the denoising process, we applied a 2D Fourier transform to the images generated at each step of the diffusion model. We then calculated the relative log magnitude of each frequency component. For an image $x$, we denote this result as $RLM(x)$. When it comes to Stable Diffusion, a latent diffusion model, we utilized a pre-trained variational autoencoder to transform the latent at each step into the image domain for frequency component analysis. The visualized results of $RLM(x_t)$ across all time step $t$ are depicted in ~\cref{fig:freq_adm,fig:freq_SD}.
As anticipated, the frequency components of the image $x_{1000}$ at $t=1000$, which represents pure Gaussian noise, exhibited a flat distribution. However, as the denoising process progressed, we observed changes in the frequency components of the images. To quantify these changes, we defined the frequency change at a specific step $t$ as the difference between the frequency components of consecutive steps: $RLM(x_{t}) - RLM(x_{t+1})$. These differences by time step $t$ are shown in ~\cref{fig:freq_change_ADM,fig:freq_change_SD}.
This approach allows us to track the evolution of frequency components throughout the denoising process. By examining these spectral changes, we can gain insights into how different frequency bands are affected at various stages of image generation.

\noindent
\textbf{Major Changes Occur in the Early and Late Stages.}
The visualization of our spectral analysis is presented in ~\cref{fig:frequency_change}. Consistent with previous works, we observe that the early stages of the denoising process (closer to $t=1000$) primarily form coarse, low-frequency components, as evident in the lower right area of ~\cref{fig:freq_change_ADM,fig:freq_change_SD}. The red regions indicating an increase in the heatmap demonstrate that the low-frequency components are formed predominantly in the early stages. Conversely, the later stages (approaching $t=0$) are dominated by changes in fine, high-frequency components, as shown in the upper left area of these graphs.
To further clarify this phenomenon, we isolated the increasing low-frequency components in the early stages and the decreasing high-frequency components in the later stages, and show their respective averages in ~\cref{fig:avg_change_adm,fig:avg_change_SD}. This separation demonstrates that low-frequency elements undergo significant changes early in the process, while high-frequency elements experience substantial modifications in the later stages. Notably, the intermediate stages show relatively minor changes in image information from a frequency perspective.




\section{Beta Sampling for Step Reduction}
\label{sec:methods}


Inspired by spectral analysis, we propose a novel sampling technique for step reduction in diffusion models. Our previous investigations revealed a significant phenomenon in the image-denoising process of diffusion models: the most critical and substantial changes are predominantly concentrated in the initial and final stages of the model's operation.
This observation suggests that when selecting a reduced number of steps compared to the total time steps used during training, there is a compelling need to develop a method that can allocate more steps to the early and late stages of the process.
Consequently, we propose a novel sampling strategy that allocates a higher density of steps to the initial and final stages of the denoising process based on Beta distribution. This method aims to concentrate computational resources on the periods where the most crucial transformations occur. 


To sample Beta distribution-like time steps in the denoising process, our method leverages the Probability Integral Transform (PIT), a well-established method for generating samples from a target distribution using uniform random variables~\cite{AngusPIT}.
We begin with a set of uniformly sampled time steps ${t_i}$, where $i = 1, 2, ..., N$, and $t_i \in [0, T-1]$. 
Then we normalize the time steps: ${t'_i} = {t_i}/T$.
These time steps are initially equidistant, following a Uniform distribution $U(0,1)$.
We choose a Beta distribution $B(\alpha, \beta)$ as our target distribution for the time steps. The Beta distribution is selected for its flexibility in modeling various shapes over a finite interval $[0, 1]$, which aligns well with the normalized time step range.
The CDF of the beta distribution, denoted as $F(x; \alpha, \beta)$, is computed.
We apply the PIT to transform our uniform samples into samples from the Beta distribution. For each uniformly sampled time step ${t'_i}$, we compute:
\begin{equation}
    t^{B}_i = F^{-1}({t'_i}; \alpha, \beta)
    \label{eq:PIT}
\end{equation}
 where $F^{-1}$ is the inverse CDF (or Percent Point Function, PPF) of the Beta distribution.
If necessary, we rescale the transformed time steps $t^{B}_i$ to ensure they span the full range of the diffusion process.
Our method does not involve random sampling from the Beta distribution. Instead, it employs distribution equalization through the PIT to achieve fixed-point sampling. This approach ensures a deterministic and consistent allocation of time steps that adheres to the desired Beta distribution, offering a more controlled and reproducible sampling process compared to random sampling methods.

By carefully selecting the $\alpha$ and $\beta$ parameters of the Beta distribution, we can precisely control the density of time steps at different stages of the denoising process. This adaptive sampling strategy enables a more efficient allocation of computational resources, focusing on the most critical phases of the diffusion process while maintaining a smooth transition throughout the entire range.

\section{Experiments}
\label{sec:exp}

\subsection{Experimental Setup}

\noindent
\textbf{Diffusion Models.}
To validate the effectiveness of our method, we conducted experiments using representative pre-trained diffusion models without retraining or fine-tuning. As in ~\cref{sec:analysis}, we employed the ADM-G~\cite{dhariwal2021diffusion_adm} trained on ImageNet 64$\times$64 and the Stable Diffusion~\cite{rombach2022ldm} for our experiments. For ADM-G, we utilized the DDIM~\cite{song2021ddim} sampler, while for Stable Diffusion, we employed PLMS~\cite{liu2022pndm_plms}. 

\noindent
\textbf{Metrics.}
We used Fréchet Inception Distance (FID)~\cite{heusel2017gans_fid} and Inception Score (IS)~\cite{salimans2016improved_IS} as our evaluation metrics, consistent with most previous works in this field. FID measures the quality difference between generated images and target images, with lower values indicating greater similarity. Higher IS values suggest that the generated images exhibit high quality and diversity.

\noindent
\textbf{Baselines.}
We conducted a comparative analysis with uniform sampling and AutoDiffusion~\cite{li2023autodiffusion}. uniform sampling is the most commonly used technique when decreasing the number of time steps. AutoDiffusion, introduced to address the limitations of uniform sampling, employs an evolutionary search algorithm to identify optimal time steps that minimize the FID. We implemented AutoDiffusion as described in the original paper, conducting a 10 epoch evolutionary search and selecting the optimal time step configuration. 

\noindent
\textbf{Implementation Details.}
In our experiments with ADM-G, we generated 50,000 samples using random class guidance. For Stable Diffusion, we produced 10,000 samples using captions from the COCO dataset. We then calculated FID and IS scores for these generated samples to assess the performance of our method. 

\begin{figure*}
  \centering
    \includegraphics[width=\linewidth]{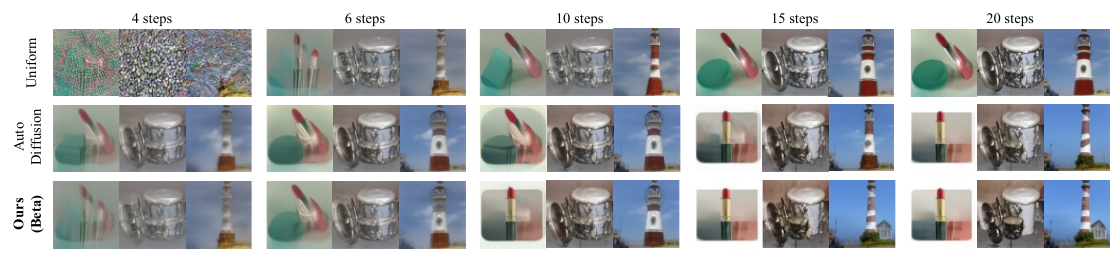}
  \caption{Examples generated by ADM-G~\cite{dhariwal2021diffusion_adm} on ImageNet 64$\times$64 with various sampling strategies.}
  \label{fig:result_adm}
\end{figure*}

\subsection{Results}
\label{sec:results}

\noindent
\textbf{Beta Sampling Outperforms Uniform Sampling and Competes with AutoDiffusion.}
\Cref{tab:result_adm} and \Cref{tab:result_sd} present a comparison of FID and IS scores for generated images using different sampling methods (Beta Sampling, AutoDiffusion, and uniform sampling) in ADM-G and Stable Diffusion (SD) models, respectively.
In both ADM and SD experiments, the three sampling methods show similar trends. 
At 50 and 100 steps, AutoDiffusion's genetic algorithm search consumes significant time, and since the step count is high enough to approach a solution-like image, the FID and IS scores for both uniform and Beta Sampling become similar.~\cite{sabour2024align}
However, at 10, 15, and 20 steps, Beta Sampling shows clear improvements over uniform sampling. 
In ADM-G, Beta Sampling even outperforms AutoDiffusion, while in Stable Diffusion, AutoDiffusion performs slightly better. 
Nevertheless, considering AutoDiffusion's long search time, Beta Sampling proves more efficient. 
In very low step settings of 4 and 6, AutoDiffusion yields better FID and IS scores than Beta Sampling
However, while uniform sampling shows a significant performance drop, Beta Sampling exhibits only marginal performance degradation. 
In summary, at higher step counts (50, 100), the difference between sampling methods is negligible, but at practical step counts (10, 15, 20), Beta Sampling excels in both quality and efficiency, whereas in near one-shot environments (4, 6 steps), AutoDiffusion performs better.


~\Cref{fig:result_adm} shows generated examples from identical initial noise. At 4 steps, Beta Sampling produces blurrier results compared to AutoDiffusion. However, at 6 and 10 steps, both Beta Sampling and AutoDiffusion yield clearer images than uniform sampling. Given the computationally intensive search process of AutoDiffusion, Beta Sampling emerges as a more efficient alternative. At 15 and 20 steps, all strategies produce clear images, with both uniform and Beta Sampling offering efficient options. Overall, except for 4 steps, Beta Sampling demonstrates competitive image quality without the additional time or computational burdens.

\begin{figure*}
  \centering
    \includegraphics[width=\linewidth]{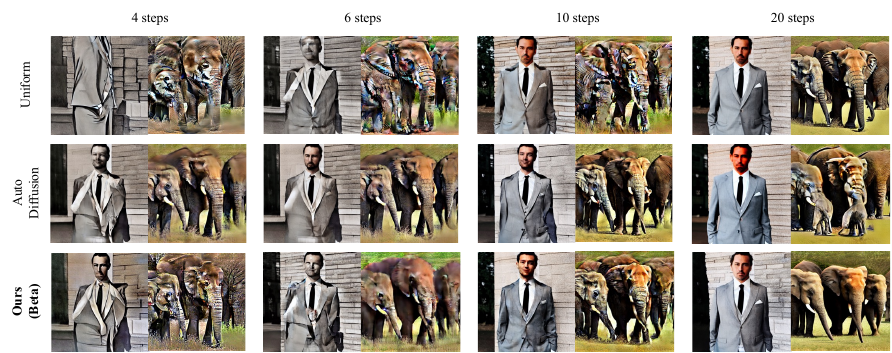}
  \caption{Examples generated by Stable Diffusion~\cite{rombach2022ldm} with various sampling strategies. The text prompts used for generation are "A man who is wearing a suit and tie" and "Two large elephants are standing beside each other".}
  \label{fig:result_sd}
\end{figure*}

The quality of examples generated with Stable Diffusion can be observed in~\cref{fig:result_sd}. At 4 and 6 steps, uniform sampling exhibits structural defects and poor color representation. Beta Sampling mitigates these issues but remains inferior to AutoDiffusion, which employs a more extensive search process for large models. At 10 steps, Beta Sampling produces samples comparable in quality to AutoDiffusion. Interestingly, at 20 steps, AutoDiffusion occasionally introduces undesired artifacts into the examples. While there is considerable fluctuation depending on the prompt, Beta Sampling demonstrates competitive efficiency in all cases except at 4 steps.

\begin{table}[]
\small
    \centering
\begin{tabular}{clrr}
\toprule
\multicolumn{1}{c}{Steps} & Sampling Strategies      & FID ($\downarrow$)            & IS ($\uparrow$)             \\
\midrule
                          & Uniform       & 138.66         & 7.09           \\
\multicolumn{1}{c}{4}     & AutoDiffusion & \textbf{17.86} & \textbf{34.88} \\
                          & \textbf{Ours (Beta)}          & 31.64          & 25.27          \\
                          \midrule
                          & Uniform       & 23.71          & 31.53          \\
\multicolumn{1}{c}{6}     & AutoDiffusion & \textbf{11.17} & \textbf{43.47} \\
                          & \textbf{Ours (Beta)}          & 13.12          & 41.30          \\
                          \midrule
                          & Uniform       & 8.86           & 46.50           \\
\multicolumn{1}{c}{10}    & AutoDiffusion & 6.24           & 57.85          \\
                          & \textbf{Ours (Beta)}          & \textbf{6.13}  & \textbf{58.15} \\
                          \midrule
                          & Uniform       & 5.38           & 54.82          \\
\multicolumn{1}{c}{15}    & AutoDiffusion & 4.92           & 64.03          \\
                          & \textbf{Ours (Beta)}          & \textbf{4.43}  & \textbf{66.28} \\
                          \midrule
                          & Uniform       & 4.35           & 58.41          \\
\multicolumn{1}{c}{20}    & AutoDiffusion & \textbf{3.93}  & 68.05          \\
                          & \textbf{Ours (Beta)}          & \textbf{3.93}  & \textbf{71.42} \\
                          \midrule
\multicolumn{1}{c}{50}    & Uniform       & 3.20            & 69.32          \\
                          & \textbf{Ours (Beta)}          & \textbf{3.17}  & \textbf{73.85} \\
                          \midrule
\multicolumn{1}{c}{100}    & Uniform       & \textbf{3.19}        & 71.50          \\
                          & \textbf{Ours (Beta)}          & \textbf{ 3.19}  & \textbf{73.85} \\
                          \bottomrule
\end{tabular}
    \caption{FID ($\downarrow$) and IS ($\uparrow$) scores for ADM-G~\cite{dhariwal2021diffusion_adm} on ImageNet 64$\times$64 with various number of time steps and sampling strategies. For Beta Sampling, we set $\alpha=\beta=0.5$ for steps from 4 to 20, and $\alpha=\beta=0.9$ for steps 50 and 100.}
    \label{tab:result_adm}
\end{table}

\begin{table}[]
\small
    \centering
\begin{tabular}{clrr}
\toprule
\multicolumn{1}{c}{Steps} & Sampling Strategies                                                       & FID ($\downarrow$) & IS ($\uparrow$) \\
\midrule
                          & Uniform                                                        & 38.22              & 16.06           \\
\multicolumn{1}{c}{4}     & AutoDiffusion & \textbf{20.18}     & \textbf{23.10}  \\
                          & \textbf{Ours (Beta)}                                                           & 34.71              & 16.95           \\
                          \midrule
                          & Uniform                                                        & 32.40              & 17.99           \\
\multicolumn{1}{c}{6}     & AutoDiffusion & \textbf{17.57}     & \textbf{23.83}  \\
                          & \textbf{Ours (Beta)}                                                           & 22.48              & 21.83           \\
                          \midrule
                          & Uniform                                                        & 19.16              & 22.04           \\
\multicolumn{1}{c}{10}    & AutoDiffusion & \textbf{13.20}     & \textbf{26.52}  \\
                          & \textbf{Ours (Beta)}                                                           & 16.45              & 25.51           \\
                          \midrule
                          & Uniform                                                        & 14.57              & 27.40           \\
\multicolumn{1}{c}{20}    & AutoDiffusion & \textbf{13.08}     & 25.21           \\
                          & \textbf{Ours (Beta)}                                                           & 15.10              & \textbf{27.46}  \\
                          \midrule
\multicolumn{1}{c}{50}    & Uniform                                                        & 15.92              & 27.76           \\
                          & \textbf{Ours (Beta)}                                                           & \textbf{15.89}               & \textbf{27.85}  \\
                          \midrule
\multicolumn{1}{c}{100}    & Uniform                                                        & 15.66             & 27.76           \\
                          & \textbf{Ours (Beta)}                                                           & \textbf{15.65}               & \textbf{27.78}  \\
                          \bottomrule
\end{tabular}
    \caption{FID ($\downarrow$) and IS ($\uparrow$) scores for Stable Diffusion~\cite{rombach2022ldm} with various number of time steps and sampling strategies. For Beta Sampling, we set $\alpha=\beta=0.6$ for steps from 4 to 20, $\alpha=\beta=0.7$ for 50 steps and $\alpha=\beta=0.9$ for 100 steps.}
    \label{tab:result_sd}
\end{table}

\begin{figure}[t]
  \centering
    \includegraphics[width=\linewidth]{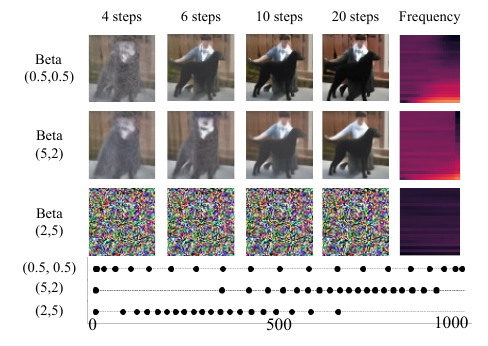}
  \caption{Examples generated by ADM-G~\cite{dhariwal2021diffusion_adm} on ImageNet 64$\times$64 using Beta Sampling with various distribution shapes. The frequency column provides a relative frequency analysis of the 20-step generation, highlighting differences in the process.}
  \label{fig:ablation_beta}
\end{figure}

\noindent
\textbf{Ablation Study (1): Symmetric Beta Sampling Performs Better.}
\cref{tab:ablation_beta} and \cref{fig:ablation_beta} present the FID and IS evaluations and sample images generated using these three Beta Sampling with different parameters. 
For Beta(2,5) with $\alpha < \beta$, which concentrates sampling in the later stages, the generated images lacked the necessary low-frequency generation, resulting in noisy and incomprehensible images, as evidenced by very poor FID and IS scores. This demonstrates that the early steps are crucial for forming low-frequency components, which need to be shaped before the high-frequency components are generated.
Beta(5,2) with $\alpha > \beta$ focused on early stages, producing images with good low-frequency content but lacking high-frequency details, resulting in blurred edges and inferior FID and IS scores compared to uniform sampling. In contrast, our proposed Beta(0.5,0.5) with $\alpha = \beta$ balanced early and late stage sampling, preserving both content and detail, and achieving the best FID and IS scores. 
These results confirm our hypothesis that the early and late steps of the denoising process are crucial for enhancing low-frequency and high-frequency components, respectively.

\begin{table}[]
\small
    \centering
\begin{tabular}{clrr}
\toprule
\multicolumn{1}{c}{Steps} & Sampling Strategies   & FID ($\downarrow$) & IS ($\uparrow$) \\
\midrule
                          & \textbf{Beta (0.5,0.5)} & \textbf{31.64}              & \textbf{25.27}           \\
\multicolumn{1}{c}{4}     & Beta (5,2)     & 52.65              & 16.89           \\
                          & Beta (2,5)     & 417.03             & 1.16            \\
                          \midrule
                          & \textbf{Beta (0.5,0.5)} & \textbf{13.12}              & \textbf{41.30}           \\
\multicolumn{1}{c}{6}     & Beta (5,2)     & 31.74              & 25.18           \\
                          & Beta (2,5)     & 419.33             & 1.18            \\
                          \midrule
                          & \textbf{Beta (0.5,0.5)} & \textbf{6.13}               & \textbf{58.15}           \\
\multicolumn{1}{c}{10}    & Beta (5,2)     & 19.82              & 34.00              \\
                          & Beta (2,5)     & 420.46             & 1.21            \\
                          \midrule
                          & \textbf{Beta (0.5,0.5)} & \textbf{4.43}               & \textbf{66.28}           \\
\multicolumn{1}{c}{15}    & Beta (5,2)     & 15.17              & 39.53           \\
                          & Beta (2,5)     & 418.37             & 1.28            \\
                          \midrule
                          & \textbf{Beta (0.5,0.5)} & \textbf{3.93}               & \textbf{71.42}           \\
\multicolumn{1}{c}{20}    & Beta (5,2)     & 13.21              & 42.31           \\
                          & Beta (2,5)     & 416.31             & 1.32         \\
                          \bottomrule
\end{tabular}
    \caption{FID ($\downarrow$) and IS ($\uparrow$) scores for ADM-G~\cite{dhariwal2021diffusion_adm} on ImageNet 64$\times$64 with Beta Sampling of various distribution shape and their corresponding sampled step distributions.}
    \label{tab:ablation_beta}
\end{table}

\begin{figure}[t]
  \centering
\begin{subfigure}{0.46\linewidth}
    \includegraphics[width=\linewidth]{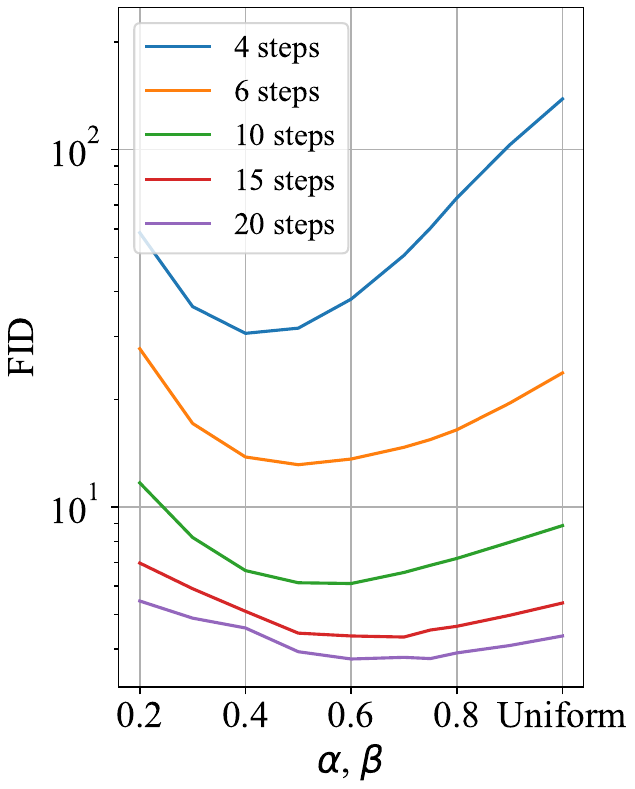}
    \caption{FID}
    \label{fig:short-a}
  \end{subfigure}
  \begin{subfigure}{0.45\linewidth}
    \includegraphics[width=\linewidth]{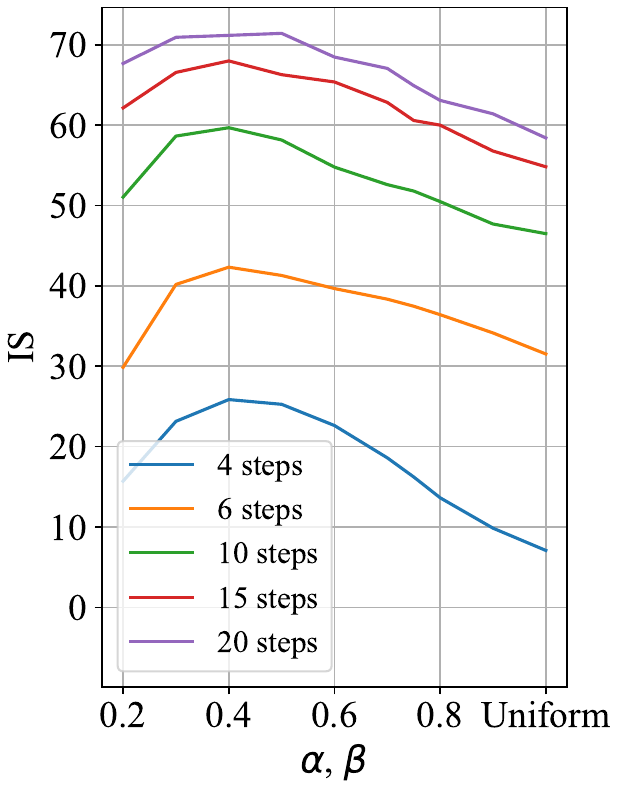}
    \caption{IS}
    \label{fig:short-b}
  \end{subfigure}
  \caption{FID ($\downarrow$) and IS ($\uparrow$) scores for ADM-G~\cite{dhariwal2021diffusion_adm} on ImageNet 64$\times$64 with Beta Sampling $Beta(\alpha,\beta)$ of various hyperparameter $\alpha=\beta$ and various time steps.}
  \label{fig:ablation_alpha}
\end{figure}

\noindent
\textbf{Ablation Study (2): Hyperparameter Analysis.}
~\Cref{fig:ablation_alpha} illustrates the changes in FID and IS performance as a function of the hyperparameter $\alpha=\beta$. When $\alpha=\beta=1$, the distribution is equivalent to a uniform distribution. The observed changes in FID and IS performance as the hyperparameter decreases from the uniform distribution demonstrate the effectiveness of Beta Sampling.
Specifically, 
we set $\alpha = \beta = 0.5$ for ADM-G, and $\alpha = \beta= 0.6$ for Stable Diffusion except for the 50- and 100-step scenarios. For this relatively large NFE region, we identified the optimal parameters individually.

\begin{figure}[t]
  \centering
    \includegraphics[width=\linewidth]{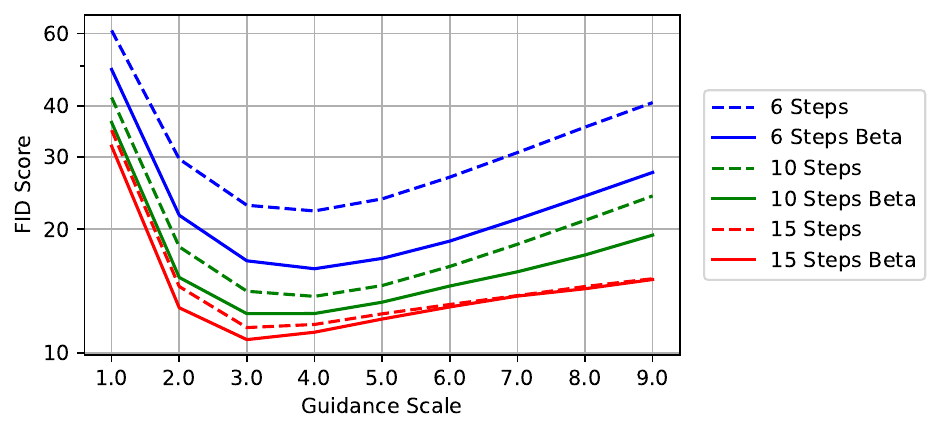}
  \caption{FID ($\downarrow$) scores for Stable Diffusion~\cite{rombach2022ldm} with Beta Sampling across various guidance scale.}
  \label{fig:guidance-scale}
\end{figure}

\noindent
\textbf{Ablation Study (3): Performance by Guidance Scale.}
~\Cref{fig:guidance-scale} shows FID scores for various guidance scales with and without Beta Sampling on Stable Diffusion. Across multiple steps, the Beta Distribution consistently reduces FID scores at all scales, demonstrating its stable performance.

\section{Discussion}
\label{sec:disc}
\begin{table}[h]
\small
    \centering
    \begin{tabular}{lcccc}
        \hline
        \textbf{Method} & \textbf{Steps} & \textbf{Search } & \textbf{Cost} & \textbf{FID} \\
        \hline
        AutoDiffusion & 10 & Genetic Algo. & 31h+ & 6.24 \\
        Ours & 10 & Param. Search & 19m & 6.10 \\
        \hline
    \end{tabular}
    \caption{Efficiency comparison with AutoDiffusion}
    \label{tab:comparison-cost}
\end{table}

\begin{figure}[h]
  \centering
  \includegraphics[width=0.9\linewidth]{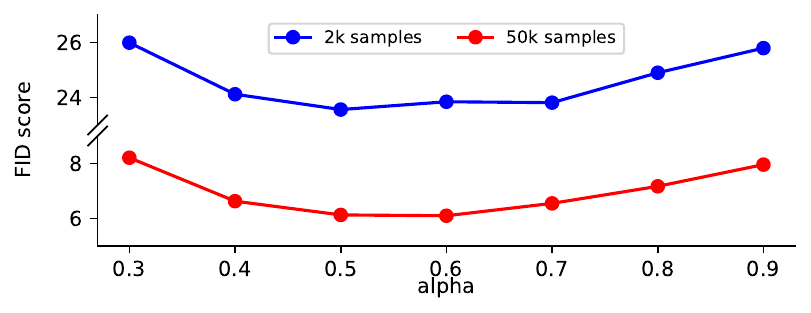}
   \caption{FID scores measured with 2k samples and 50k samples across $\alpha=\beta$ of Beta distribution.}
   \label{fig:fid-2k50k}
\end{figure}

\noindent
\textbf{Beta Sampling is Highly Efficient.}
One key feature of Beta Sampling is that it is both training-free and highly efficient. Various denoising speed-up techniques typically rely on either distillation from existing models or require additional training. For instance, in the case of InstaFlow~\cite{liu2023instaflow}, the process takes 199 days on an A100. In contrast, both AutoDiffusion and our method are training-free, search-based approaches, but differ significantly in efficiency. \Cref{tab:comparison-cost} demonstrates the superior efficiency of Beta Sampling compared to AutoDiffusion. For the ADM-G 10-step setting, AutoDiffusion requires a minimum of 31 hours for the search, while our method takes only 19 minutes with a single A100. As shown in \cref{fig:fid-2k50k}, by sampling just 2k images for parameter search, we were able to effectively determine the optimal $\alpha$. This process includes denoising sample generation, and the gap in efficiency becomes even more pronounced for high-resolution models or when a higher number of steps is involved.

\begin{figure}[t]
  \centering
\begin{subfigure}{0.40\linewidth}
    \includegraphics[width=\linewidth]{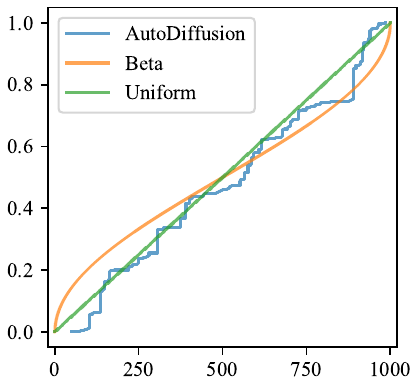}
    \caption{4 steps.}
    \label{fig:short-a}
  \end{subfigure}
  \begin{subfigure}{0.40\linewidth}
    \includegraphics[width=\linewidth]{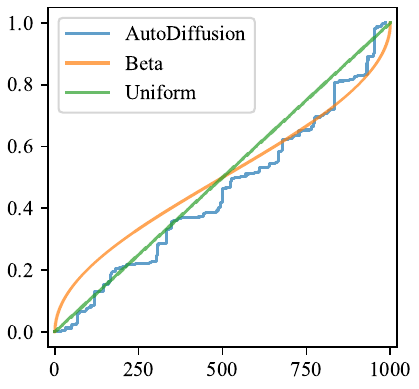}
    \caption{6 steps.}
    \label{fig:short-b}
  \end{subfigure}
\begin{subfigure}{0.4\linewidth}
    \includegraphics[width=\linewidth]{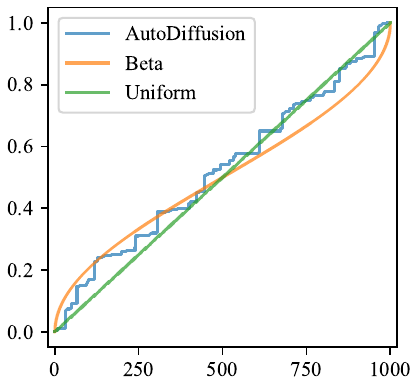}
    \caption{10 steps.}
    \label{fig:short-a}
  \end{subfigure}
  \begin{subfigure}{0.4\linewidth}
    \includegraphics[width=\linewidth]{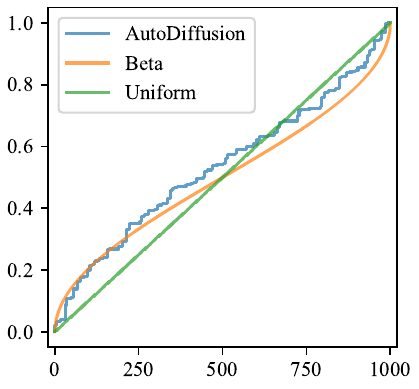}
    \caption{15 steps.}
    \label{fig:short-b}
  \end{subfigure}
  \caption{A cumulative histogram of occurrence number of time steps sampled by AutoDiffusion~\cite{li2023autodiffusion} on ADM-G~\cite{dhariwal2021diffusion_adm} with various number of time steps. To enhance the visibility of the histogram, the time steps were sampled 150 times for 4 and 6 steps, and 100 times for 10 and 15 steps, respectively.}
  \label{fig:step_cumul_hist}
\end{figure}

\noindent
\textbf{AutoDiffusion Behaves Like Beta Sampling.}
Our experiments revealed that Beta Sampling achieves results comparable to AutoDiffusion when using 10 or more time steps, but underperforms with extremely small time steps, such as 4 or 6 steps. 
To better understand these differences, we analyzed the distribution of time steps extracted from AutoDiffusion by examining the cumulative histogram of step frequencies, as shown in Fig.~\ref{fig:step_cumul_hist}. 
For enhanced visibility, we repeated step sampling 150 times for 4 and 6 steps, and 50 times for 10 and 15 steps. 
The histogram revealed that for 4 and 6 steps, the distribution was nearly uniform, whereas for 10 and 15 steps, the distribution resembled a Beta distribution.
This indicates that AutoDiffusion effectively operates similarly to Beta Sampling when using 10 or more steps.
Consequently, for scenarios requiring 10 or more time steps, Beta Sampling emerges as a more efficient choice compared to AutoDiffusion, which involves a longer searching time.

\noindent
\textbf{ADM-G vs. Stable Diffusion.}
We validated the effectiveness of Beta Sampling using two diffusion models, ADM-G and Stable Diffusion. 
Our results demonstrated that Beta Sampling is more advantageous than uniform sampling for both models and is particularly more efficient than AutoDiffusion for ADM-G with steps greater than 10. 
However, in the case of Stable Diffusion, Beta Sampling showed marginally inferior results to AutoDiffusion even at 10 steps. 
This discrepancy can be attributed to the differences in the spectral analysis of ADM-G and Stable Diffusion. 
As depicted in Fig.~\ref{fig:frequency_change}, ADM's frequency changes exhibit peaks in both high and low frequencies concentrated at the extremities of the steps with a steep decline in variance. 
In contrast, Stable Diffusion's frequency changes show a high-frequency peak around the 100-step mark, with low-frequency changes tapering off more gradually. 
Therefore, while Beta Sampling is well-suited to ADM's frequency change distribution, it is not entirely compatible with Stable Diffusion's distribution. 
Future improvements in generation performance for Stable Diffusion could be achieved by identifying hyperparameters or distributions that better align with its frequency change characteristics.

\noindent
\textbf{Limitations and Future Works.}
One limitation is that our approach relies on pre-determined spectral analyses, which may not fully capture the dynamic nature of the diffusion process across different datasets and model architectures. 
Future work could address these limitations by adaptively sampling the most critical steps to further enhance the efficiency and generalizability of our method.
\section{Conclusion}

In this paper, we introduced an efficient time step sampling method for generative diffusion models, leveraging frequency domain analysis to optimize the image generation process. Our approach replaces traditional uniform sampling with a Beta distribution-like method, emphasizing critical steps in the early and late stages of diffusion. We validated our hypothesis—that significant changes in image content occur at specific steps—through Fourier transform analysis, revealing substantial low-frequency changes early on and high-frequency adjustments later. Experiments with ADM-G and Stable Diffusion showed our method consistently outperforms uniform sampling, achieving better FID and IS scores, and offers competitive efficiency compared to state-of-the-art techniques like AutoDiffusion. This work provides a practical framework for enhancing diffusion model efficiency by focusing computational resources on the most impactful steps, with potential for further optimization and adaptive techniques in future research.

\section*{Acknowledgements}
This work was partly supported by
Basic Science Research Program through the National Research Foundation of Korea (NRF) funded by the Ministry of Education (2022R1I1A1A01071970)
and
Institute of Information \& communications Technology Planning \& Evaluation (IITP) grant funded by the Korea government(MSIT) (No.RS-2021-II212068, Artificial Intelligence Innovation Hub).

\clearpage
{\small
\bibliographystyle{ieee_fullname}
\bibliography{main}

@String(CVPR  = {IEEE Conf. Comput. Vis. Pattern Recog.})

@String(ICCV  = {Int. Conf. Comput. Vis.})

@String(AAAI  = {AAAI})

@String(CVPR  = {CVPR})

@String(ICCV  = {ICCV})

@inproceedings{park2023understanding,
 author = {Park, Yong-Hyun and Kwon, Mingi and Choi, Jaewoong and Jo, Junghyo and Uh, Youngjung},
 booktitle = {Advances in Neural Information Processing Systems},
 editor = {A. Oh and T. Neumann and A. Globerson and K. Saenko and M. Hardt and S. Levine},
 pages = {24129--24142},
 publisher = {Curran Associates, Inc.},
 title = {Understanding the Latent Space of Diffusion Models through the Lens of Riemannian Geometry},
 url = {https://proceedings.neurips.cc/paper_files/paper/2023/file/4bfcebedf7a2967c410b64670f27f904-Paper-Conference.pdf},
 volume = {36},
 year = {2023}
}

@InProceedings{li2023autodiffusion,
    author    = {Li, Lijiang and Li, Huixia and Zheng, Xiawu and Wu, Jie and Xiao, Xuefeng and Wang, Rui and Zheng, Min and Pan, Xin and Chao, Fei and Ji, Rongrong},
    title     = {AutoDiffusion: Training-Free Optimization of Time Steps and Architectures for Automated Diffusion Model Acceleration},
    booktitle = {Proceedings of the IEEE/CVF International Conference on Computer Vision (ICCV)},
    month     = {October},
    year      = {2023},
    pages     = {7105-7114}
}

@article{xu2019frequency,
  title={Frequency principle: Fourier analysis sheds light on deep neural networks},
  author={Xu, Zhi-Qin John and Zhang, Yaoyu and Luo, Tao and Xiao, Yanyang and Ma, Zheng},
  journal={arXiv preprint arXiv:1901.06523},
  year={2019}
}

@InProceedings{xu2019training,
author="Xu, Zhi-Qin John
and Zhang, Yaoyu
and Xiao, Yanyang",
editor="Gedeon, Tom
and Wong, Kok Wai
and Lee, Minho",
title="Training Behavior of Deep Neural Network in Frequency Domain",
booktitle="Neural Information Processing",
year="2019",
publisher="Springer International Publishing",
address="Cham",
pages="264--274",
isbn="978-3-030-36708-4"
}

@article{rahaman2018spectral,
  title={On the spectral bias of deep neural networks},
  author={Rahaman, Nasim and Arpit, Devansh and Baratin, Aristide and Draxler, Felix and Lin, Min and Hamprecht, Fred A and Bengio, Yoshua and Courville, Aaron},
  journal={arXiv preprint arXiv:1806.08734},
  volume={4},
  year={2018}
}

@inproceedings{schwarz2021frequency,
 author = {Schwarz, Katja and Liao, Yiyi and Geiger, Andreas},
 booktitle = {Advances in Neural Information Processing Systems},
 editor = {M. Ranzato and A. Beygelzimer and Y. Dauphin and P.S. Liang and J. Wortman Vaughan},
 pages = {18126--18136},
 publisher = {Curran Associates, Inc.},
 title = {On the Frequency Bias of Generative Models},
 url = {https://proceedings.neurips.cc/paper_files/paper/2021/file/96bf57c6ff19504ff145e2a32991ea96-Paper.pdf},
 volume = {34},
 year = {2021}
}

@article{khayatkhoei2022spatial, title={Spatial Frequency Bias in Convolutional Generative Adversarial Networks}, volume={36}, url={https://ojs.aaai.org/index.php/AAAI/article/view/20675}, DOI={10.1609/aaai.v36i7.20675}, abstractNote={Understanding the capability of Generative Adversarial Networks (GANs) in learning the full spectrum of spatial frequencies, that is, beyond the low-frequency dominant spectrum of natural images, is critical for assessing the reliability of GAN-generated data in any detail-sensitive application. In this work, we show that the ability of convolutional GANs to learn an image distribution depends on the spatial frequency of the underlying carrier signal, that is, they have a bias against learning high spatial frequencies. Our findings are consistent with the recent observations of high-frequency artifacts in GAN-generated images, but further suggest that such artifacts are the consequence of an underlying bias. We also provide a theoretical explanation for this bias as the manifestation of linear dependencies present in the spectrum of filters of a typical generative Convolutional Neural Network (CNN). Finally, by proposing a proof-of-concept method that can effectively manipulate this bias towards other spatial frequencies, we show that the bias is not fixed and can be exploited to explicitly direct computational resources towards any specific spatial frequency of interest in a dataset, with minimal computational overhead.}, number={7}, journal={Proceedings of the AAAI Conference on Artificial Intelligence}, author={Khayatkhoei, Mahyar and Elgammal, Ahmed}, year={2022}, month={Jun.}, pages={7152-7159} }

@InProceedings{frank2020leveraging,
  title = 	 {Leveraging Frequency Analysis for Deep Fake Image Recognition},
  author =       {Frank, Joel and Eisenhofer, Thorsten and Sch{\"o}nherr, Lea and Fischer, Asja and Kolossa, Dorothea and Holz, Thorsten},
  booktitle = 	 {Proceedings of the 37th International Conference on Machine Learning},
  pages = 	 {3247--3258},
  year = 	 {2020},
  editor = 	 {III, Hal Daumé and Singh, Aarti},
  volume = 	 {119},
  series = 	 {Proceedings of Machine Learning Research},
  month = 	 {13--18 Jul},
  publisher =    {PMLR},
  url = 	 {https://proceedings.mlr.press/v119/frank20a.html}
}

@InProceedings{choi2022perception,
    author    = {Choi, Jooyoung and Lee, Jungbeom and Shin, Chaehun and Kim, Sungwon and Kim, Hyunwoo and Yoon, Sungroh},
    title     = {Perception Prioritized Training of Diffusion Models},
    booktitle = {Proceedings of the IEEE/CVF Conference on Computer Vision and Pattern Recognition (CVPR)},
    month     = {June},
    year      = {2022},
    pages     = {11472-11481}
}

@article{si2023freeu,
  title={Freeu: Free lunch in diffusion u-net},
  author={Si, Chenyang and Huang, Ziqi and Jiang, Yuming and Liu, Ziwei},
  journal={arXiv preprint arXiv:2309.11497},
  year={2023}
}

@InProceedings{rombach2022ldm,
    author    = {Rombach, Robin and Blattmann, Andreas and Lorenz, Dominik and Esser, Patrick and Ommer, Bj\"orn},
    title     = {High-Resolution Image Synthesis With Latent Diffusion Models},
    booktitle = {Proceedings of the IEEE/CVF Conference on Computer Vision and Pattern Recognition (CVPR)},
    month     = {June},
    year      = {2022},
    pages     = {10684-10695}
}

@inproceedings{ho2020ddpm,
 author = {Ho, Jonathan and Jain, Ajay and Abbeel, Pieter},
 booktitle = {Advances in Neural Information Processing Systems},
 editor = {H. Larochelle and M. Ranzato and R. Hadsell and M.F. Balcan and H. Lin},
 pages = {6840--6851},
 publisher = {Curran Associates, Inc.},
 title = {Denoising Diffusion Probabilistic Models},
 url = {https://proceedings.neurips.cc/paper_files/paper/2020/file/4c5bcfec8584af0d967f1ab10179ca4b-Paper.pdf},
 volume = {33},
 year = {2020}
}

@inproceedings{
song2021ddim,
title={Denoising Diffusion Implicit Models},
author={Jiaming Song and Chenlin Meng and Stefano Ermon},
booktitle={International Conference on Learning Representations},
year={2021},
url={https://openreview.net/forum?id=St1giarCHLP}
}

@inproceedings{
liu2022pndm_plms,
title={Pseudo Numerical Methods for Diffusion Models on Manifolds},
author={Luping Liu and Yi Ren and Zhijie Lin and Zhou Zhao},
booktitle={International Conference on Learning Representations},
year={2022},
url={https://openreview.net/forum?id=PlKWVd2yBkY}
}

@inproceedings{lu2022dpmsolver,
 author = {Lu, Cheng and Zhou, Yuhao and Bao, Fan and Chen, Jianfei and LI, Chongxuan and Zhu, Jun},
 booktitle = {Advances in Neural Information Processing Systems},
 editor = {S. Koyejo and S. Mohamed and A. Agarwal and D. Belgrave and K. Cho and A. Oh},
 pages = {5775--5787},
 publisher = {Curran Associates, Inc.},
 title = {DPM-Solver: A Fast ODE Solver for Diffusion Probabilistic Model Sampling in Around 10 Steps},
 url = {https://proceedings.neurips.cc/paper_files/paper/2022/file/260a14acce2a89dad36adc8eefe7c59e-Paper-Conference.pdf},
 volume = {35},
 year = {2022}
}

@inproceedings{
salimans2022progressive,
title={Progressive Distillation for Fast Sampling of Diffusion Models},
author={Tim Salimans and Jonathan Ho},
booktitle={International Conference on Learning Representations},
year={2022},
url={https://openreview.net/forum?id=TIdIXIpzhoI}
}

@misc{berthelot2023tract,
      title={TRACT: Denoising Diffusion Models with Transitive Closure Time-Distillation}, 
      author={David Berthelot and Arnaud Autef and Jierui Lin and Dian Ang Yap and Shuangfei Zhai and Siyuan Hu and Daniel Zheng and Walter Talbott and Eric Gu},
      year={2023},
      eprint={2303.04248},
      archivePrefix={arXiv},
      primaryClass={cs.LG}
}

@InProceedings{meng2023distillation,
    author    = {Meng, Chenlin and Rombach, Robin and Gao, Ruiqi and Kingma, Diederik and Ermon, Stefano and Ho, Jonathan and Salimans, Tim},
    title     = {On Distillation of Guided Diffusion Models},
    booktitle = {Proceedings of the IEEE/CVF Conference on Computer Vision and Pattern Recognition (CVPR)},
    month     = {June},
    year      = {2023},
    pages     = {14297-14306}
}

@misc{balaji2023ediffi,
      title={eDiff-I: Text-to-Image Diffusion Models with an Ensemble of Expert Denoisers}, 
      author={Yogesh Balaji and Seungjun Nah and Xun Huang and Arash Vahdat and Jiaming Song and Qinsheng Zhang and Karsten Kreis and Miika Aittala and Timo Aila and Samuli Laine and Bryan Catanzaro and Tero Karras and Ming-Yu Liu},
      year={2023},
      eprint={2211.01324},
      archivePrefix={arXiv},
      primaryClass={cs.CV}
}

@inproceedings{
watson2022ddss,
title={Learning Fast Samplers for Diffusion Models by Differentiating Through Sample Quality},
author={Daniel Watson and William Chan and Jonathan Ho and Mohammad Norouzi},
booktitle={International Conference on Learning Representations},
year={2022},
url={https://openreview.net/forum?id=VFBjuF8HEp}
}

@inproceedings{fang2023diffpruning,
 author = {Fang, Gongfan and Ma, Xinyin and Wang, Xinchao},
 booktitle = {Advances in Neural Information Processing Systems},
 editor = {A. Oh and T. Neumann and A. Globerson and K. Saenko and M. Hardt and S. Levine},
 pages = {16716--16728},
 publisher = {Curran Associates, Inc.},
 title = {Structural Pruning for Diffusion Models},
 url = {https://proceedings.neurips.cc/paper_files/paper/2023/file/35c1d69d23bb5dd6b9abcd68be005d5c-Paper-Conference.pdf},
 volume = {36},
 year = {2023}
}

@inproceedings{
yang2024ddsm,
title={Denoising Diffusion Step-aware Models},
author={Shuai Yang and Yukang Chen and Luozhou WANG and Shu Liu and Ying-Cong Chen},
booktitle={The Twelfth International Conference on Learning Representations},
year={2024},
url={https://openreview.net/forum?id=c43FGk8Pcg}
}

@misc{lee2023meme,
      title={Multi-Architecture Multi-Expert Diffusion Models}, 
      author={Yunsung Lee and Jin-Young Kim and Hyojun Go and Myeongho Jeong and Shinhyeok Oh and Seungtaek Choi},
      year={2023},
      eprint={2306.04990},
      archivePrefix={arXiv},
      primaryClass={cs.CV}
}

@InProceedings{yang2022spectral_diffusion,
    author    = {Yang, Xingyi and Zhou, Daquan and Feng, Jiashi and Wang, Xinchao},
    title     = {Diffusion Probabilistic Model Made Slim},
    booktitle = {Proceedings of the IEEE/CVF Conference on Computer Vision and Pattern Recognition (CVPR)},
    month     = {June},
    year      = {2023},
    pages     = {22552-22562}
}

@article{AngusPIT,
 ISSN = {00361445, 10957200},
 URL = {http://www.jstor.org/stable/2132726},
 author = {John E. Angus},
 journal = {SIAM Review},
 number = {4},
 pages = {652--654},
 publisher = {Society for Industrial and Applied Mathematics},
 title = {The Probability Integral Transform and Related Results},
 urldate = {2024-07-11},
 volume = {36},
 year = {1994}
}

@article{park2024explaining,
  title={Explaining generative diffusion models via visual analysis for interpretable decision-making process},
  author={Park, Ji-Hoon and Ju, Yeong-Joon and Lee, Seong-Whan},
  journal={Expert Systems with Applications},
  volume={248},
  pages={123231},
  year={2024},
  publisher={Elsevier}
}

@article{deja2022analyzing,
  title={On analyzing generative and denoising capabilities of diffusion-based deep generative models},
  author={Deja, Kamil and Kuzina, Anna and Trzcinski, Tomasz and Tomczak, Jakub},
  journal={Advances in Neural Information Processing Systems},
  volume={35},
  pages={26218--26229},
  year={2022}
}

@article{dhariwal2021diffusion_adm,
  title={Diffusion models beat gans on image synthesis},
  author={Dhariwal, Prafulla and Nichol, Alexander},
  journal={Advances in neural information processing systems},
  volume={34},
  pages={8780--8794},
  year={2021}
}

@INPROCEEDINGS{imagenet,
  author={Deng, Jia and Dong, Wei and Socher, Richard and Li, Li-Jia and Kai Li and Li Fei-Fei},
  booktitle={2009 IEEE Conference on Computer Vision and Pattern Recognition}, 
  title={ImageNet: A large-scale hierarchical image database}, 
  year={2009},
  volume={},
  number={},
  pages={248-255},
  doi={10.1109/CVPR.2009.5206848}}

@inproceedings{lin2014microsoft_coco,
  title={Microsoft coco: Common objects in context},
  author={Lin, Tsung-Yi and Maire, Michael and Belongie, Serge and Hays, James and Perona, Pietro and Ramanan, Deva and Doll{\'a}r, Piotr and Zitnick, C Lawrence},
  booktitle={Computer Vision--ECCV 2014: 13th European Conference, Zurich, Switzerland, September 6-12, 2014, Proceedings, Part V 13},
  pages={740--755},
  year={2014},
  organization={Springer}
}

@article{schuhmann2022laion,
  title={Laion-5b: An open large-scale dataset for training next generation image-text models},
  author={Schuhmann, Christoph and Beaumont, Romain and Vencu, Richard and Gordon, Cade and Wightman, Ross and Cherti, Mehdi and Coombes, Theo and Katta, Aarush and Mullis, Clayton and Wortsman, Mitchell and others},
  journal={Advances in Neural Information Processing Systems},
  volume={35},
  pages={25278--25294},
  year={2022}
}

@article{heusel2017gans_fid,
  title={Gans trained by a two time-scale update rule converge to a local nash equilibrium},
  author={Heusel, Martin and Ramsauer, Hubert and Unterthiner, Thomas and Nessler, Bernhard and Hochreiter, Sepp},
  journal={Advances in neural information processing systems},
  volume={30},
  year={2017}
}

@article{salimans2016improved_IS,
  title={Improved techniques for training gans},
  author={Salimans, Tim and Goodfellow, Ian and Zaremba, Wojciech and Cheung, Vicki and Radford, Alec and Chen, Xi},
  journal={Advances in neural information processing systems},
  volume={29},
  year={2016}
}

@inproceedings{liu2023instaflow,
  title={Instaflow: One step is enough for high-quality diffusion-based text-to-image generation},
  author={Liu, Xingchao and Zhang, Xiwen and Ma, Jianzhu and Peng, Jian and others},
  booktitle={The Twelfth International Conference on Learning Representations},
  year={2023}
}

@article{sabour2024align,
  title={Align your steps: Optimizing sampling schedules in diffusion models},
  author={Sabour, Amirmojtaba and Fidler, Sanja and Kreis, Karsten},
  journal={arXiv preprint arXiv:2404.14507},
  year={2024}
}
}

\clearpage
\twocolumn[

\begin{center}
    {\LARGE \bfseries Supplementary Materials\par}
    \vspace{0.5cm}
    {\large Beta Sampling is All You Need: Efficient Image Generation Strategy for\\ Diffusion Models using Stepwise Spectral Analysis\par}
    \vspace{1cm}
\end{center}

]
\appendix






\section{Experiment Details and Additional Step Analysis of ADM-G}

For our experiments with ADM-G on ImageNet 64 datasets, we utilized the official codebase~\footnote{https://github.com/openai/guided-diffusion} and the publicly released checkpoint. We generated 50,000 images and employed pre-computed sample batches from the reference datasets available in the ADM codebase to calculate the FID scores presented in Tables 1 and 3 of our main manuscript. 

\begin{figure}
  \centering
  \begin{subfigure}{0.49\linewidth}
    \includegraphics[width=\linewidth]{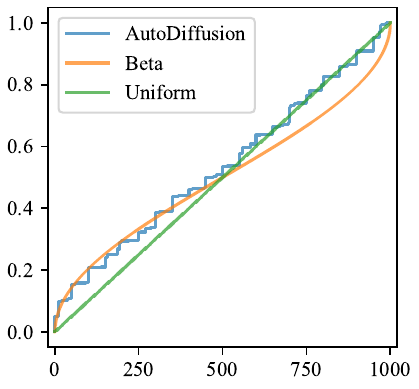}
    \caption{20 steps.}
    \label{fig:short-a}
  \end{subfigure}
  \caption{A cumulative histogram of occurrence number of time steps sampled by AutoDiffusion on ADM-G with 20 time steps.}
  \label{fig:step_cumul_hist_adm20}
\end{figure}

\Cref{fig:step_cumul_hist_adm20} illustrates the cumulative histogram of occurrence numbers for 20 steps, following the same procedure outlined in Figure 8 of the main manuscript. To enhance the visibility of the histogram, the time steps were sampled 50 times. Notably, the histogram for AutoDiffusion exhibited characteristics similar to a Beta distribution, consistent with our observations for 10 and 15 steps. This consistency across different step numbers further supports the robustness of our proposed Beta Sampling approach.

\section{Experiment Details and Step Analysis of Stable Diffusion}

In our experiments, we employed the official codebase~\footnote{https://github.com/CompVis/stable-diffusion} and the ``sd-v1-4.ckpt" checkpoint. For FID and IS measurements, we used the validation set of the COCO 2014 dataset. Unless otherwise specified, the Stable Diffusion experiments were conducted using the PLMS solver.

\begin{figure}
  \centering
  \begin{subfigure}{0.49\linewidth}
    \includegraphics[width=\linewidth]{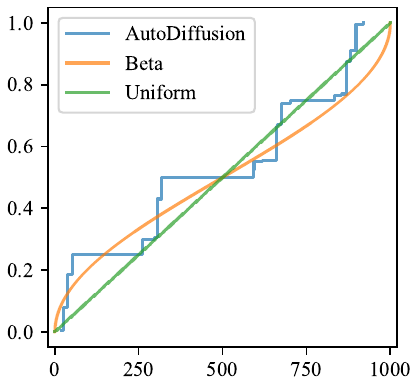}
    \caption{4 steps.}
  \end{subfigure}
  \begin{subfigure}{0.49\linewidth}
    \includegraphics[width=\linewidth]{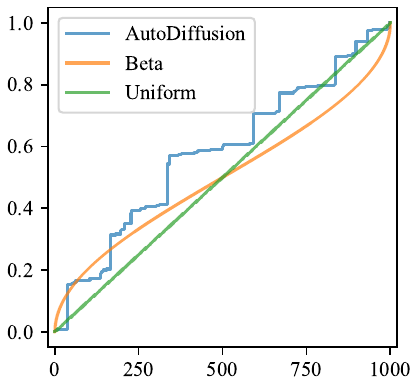}
    \caption{6 steps.}
  \end{subfigure}
  \begin{subfigure}{0.49\linewidth}
    \includegraphics[width=\linewidth]{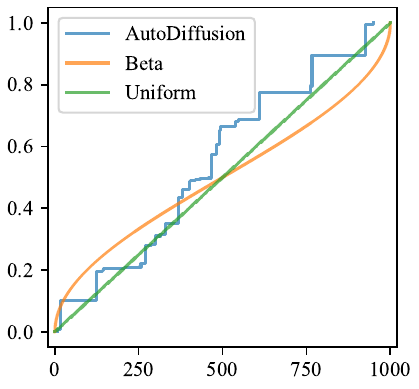}
    \caption{10 steps.}
  \end{subfigure}
  \begin{subfigure}{0.49\linewidth}
    \includegraphics[width=\linewidth]{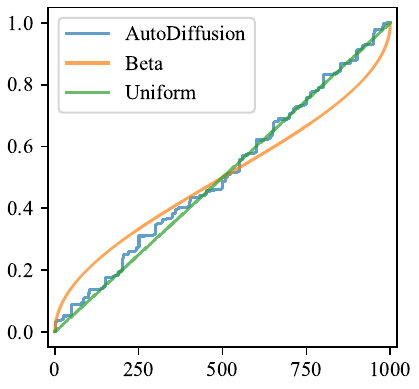}
    \caption{20 steps.}
  \end{subfigure}
  \caption{A cumulative histogram of occurrence number of time steps sampled by AutoDiffusion on Stable Diffusion with various number of time steps.}
  \label{fig:step_cumul_hist_sd}
\end{figure}

\Cref{fig:step_cumul_hist_sd} shows the Stable Diffusion version of the cumulative histogram of the occurrence number of time steps plot, which was originally presented in Fig. 8 of the main manuscript. We conducted 50 sampling iterations for each time step and observed that Beta Sampling in Stable Diffusion exhibits trends similar to AutoDiffusion, although less pronounced than in ADM-G. This difference in intensity is hypothesized to be the underlying cause of the performance difference observed in Stable Diffusion. Despite these intensity variations, the consistent emergence of Beta-like patterns across different architectures suggests the fundamental validity of our sampling approach.

\clearpage

\begin{figure*}
  \centering
    \includegraphics[width=\linewidth]{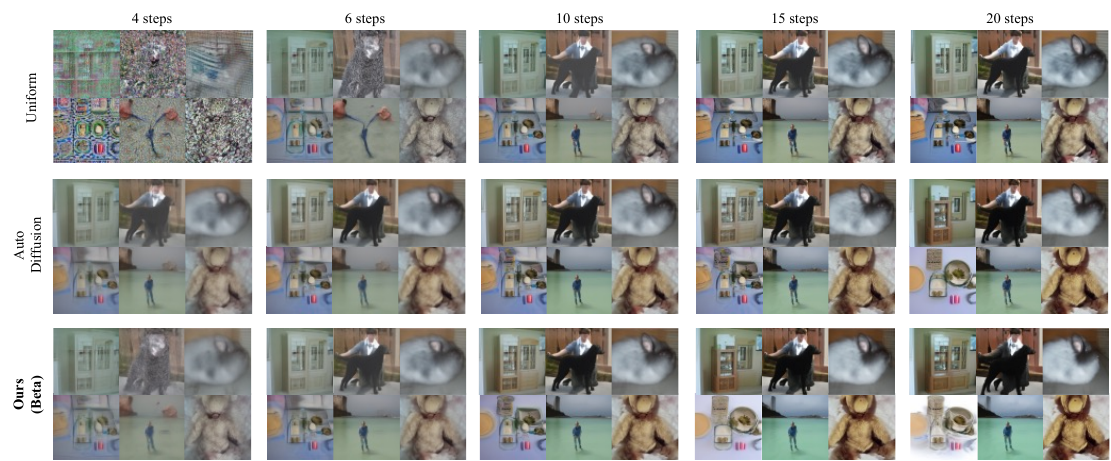}
  \caption{Examples generated by ADM-G on ImageNet 64$\times$64 with various sampling strategies.}
  \label{fig:result_adm_sup}
\end{figure*}

\begin{figure*}
  \centering
    \includegraphics[width=\linewidth]{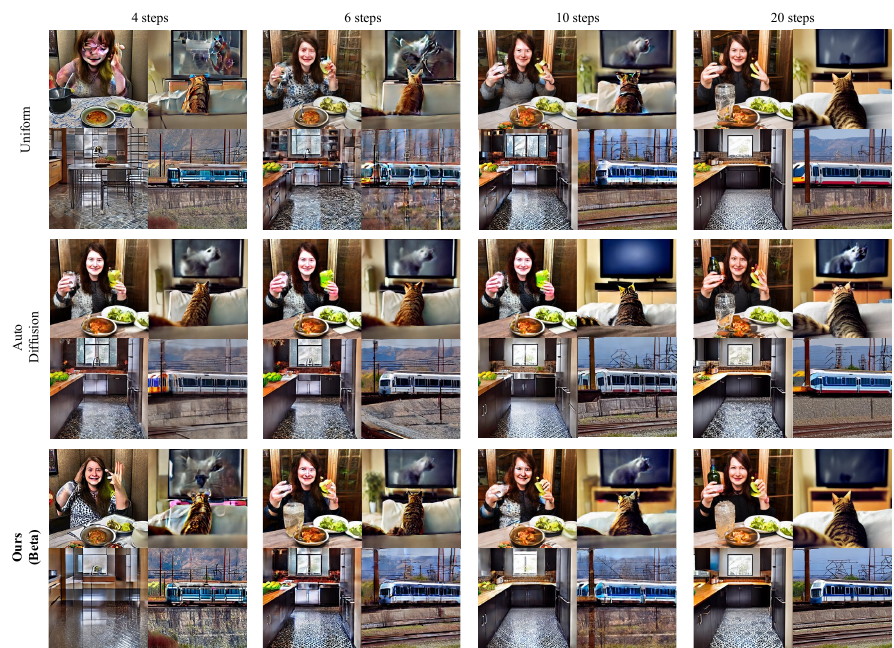}
  \caption{Examples generated by Stable Diffusion with various sampling strategies. The text prompts used for generation are ``A lady sitting at a table with food and drink, holding up two fingers", ``A cat watching TV while laying in bed", ``A kitchen with a tile floor and a metallic sink" and ``The transit train stretches down the track under the power lines".}
  \label{fig:result_sd_sup}
\end{figure*}

\clearpage



\section{Additional Examples Generated by ADM-G}
~\Cref{fig:result_adm_sup} presents generated samples from ADM-G using identical initial noise across various sampling strategies. Our comprehensive analysis reveals distinct performance patterns across different step counts. At 4 steps, Beta Sampling demonstrates markedly superior results compared to uniform sampling, which exhibits significant difficulties in maintaining structural coherence and image fidelity. While Beta Sampling at this step count does not fully match AutoDiffusion's clarity and detail preservation, it achieves a favorable balance between quality and computational efficiency. At 6 and 10 steps, both Beta Sampling and AutoDiffusion produce notably sharper and more detailed images compared to uniform sampling, with Beta Sampling achieving particularly impressive results in maintaining global structure and local details. The computational advantage of Beta Sampling becomes especially apparent when considering AutoDiffusion's intensive search process, which requires substantial additional resources without proportional quality improvements.
\section{Additional Examples Generated by Stable Diffusion}
The supplementary results for Stable Diffusion, as shown in~\cref{fig:result_sd_sup}, provide valuable insights into the performance characteristics of different sampling strategies across various step counts. At 4 and 6 steps, uniform sampling demonstrates severe limitations, manifesting as significant structural defects, inconsistent object boundaries, and poor color reproduction that substantially impact the visual quality of the generated images. Beta Sampling successfully addresses many of these shortcomings, showing marked improvements in structural coherence and color fidelity, though it does not completely match the quality level achieved by AutoDiffusion when working with larger models that benefit from its more exhaustive search process. As we increase to 10 and 20 steps, Beta Sampling demonstrates particularly impressive performance, achieving sample quality that is virtually indistinguishable from AutoDiffusion while maintaining its computational efficiency advantage. This quality parity at higher step counts is especially noteworthy given the substantial computational savings offered by Beta Sampling. Throughout our experiments across all step counts, Beta Sampling consistently emerges as a highly competitive approach, delivering superior image quality compared to uniform sampling while avoiding the computational overhead associated with AutoDiffusion's search-based methodology. These results strongly suggest that Beta Sampling represents an optimal balance between generation quality and computational efficiency in practical applications.


\section{Beta Distribution by Different Parameters}
\begin{figure}[t]
  \centering
\begin{subfigure}{0.4\linewidth}
    \includegraphics[width=\linewidth]{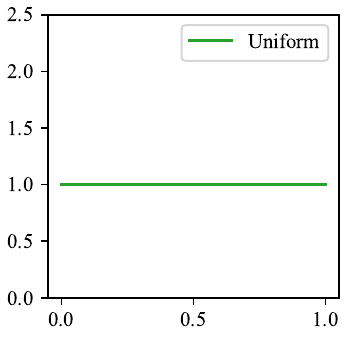}
    \caption{PDF of uniform distribution}
    \label{fig:pdf_uni}
  \end{subfigure}
  \begin{subfigure}{0.4\linewidth}
    \includegraphics[width=\linewidth]{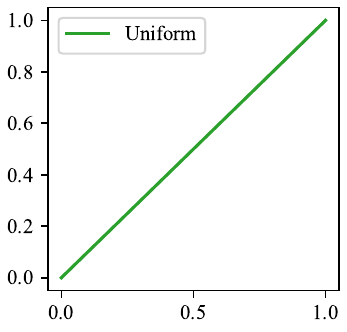}
    \caption{CDF of uniform distribution}
    \label{fig:cdf_uni}
  \end{subfigure}
  \begin{subfigure}{0.4\linewidth}
    \includegraphics[width=\linewidth]{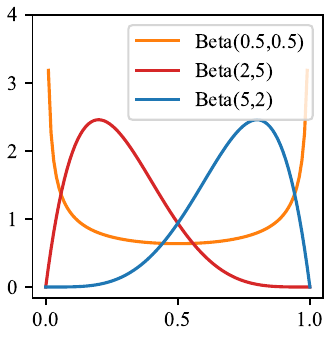}
    \caption{PDF of Beta distribution}
    \label{fig:pdf_beta}
  \end{subfigure}
  \begin{subfigure}{0.4\linewidth}
    \includegraphics[width=\linewidth]{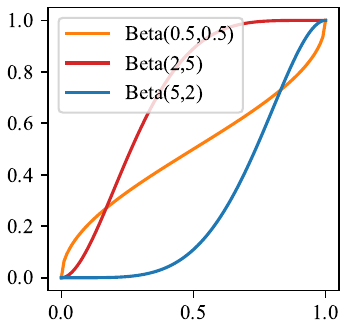}
    \caption{CDF of Beta distribution}
    \label{fig:cdf_beta}
  \end{subfigure}
  \caption{Probability Density Function (PDF) and Cumulative Distribution Function (CDF) of uniform and Beta distributions.}
  \label{fig:P-distribution}
\end{figure}

\Cref{fig:P-distribution} illustrates the Probability Density Functions (PDFs) and Cumulative Distribution Functions (CDFs) of uniform and various Beta distributions. 
The Beta distribution is characterized by two hyperparameters, $\alpha$ and $\beta$, which determine the shape of the distribution. When $\alpha > \beta$, the distribution skews to the right, meaning that sampling the denoising process based on this distribution will focus on the changes in low-frequency components during the early stages. 
Conversely, if $\alpha < \beta$, the distribution skews to the left, concentrating the denoising process on high-frequency component changes in the later stages.
Finally, when $\alpha = \beta$, the Beta distribution forms a shape with peaks at both ends, which means that sampling according to this distribution will evenly concentrate on both low-frequency changes in the early stages and high-frequency changes in the later stages. 
In our proposed method, we use a Beta distribution where $\alpha = \beta$ to ensure a balanced focus on both low-frequency and high-frequency changes throughout the denoising process. 
This approach ensures that the denoising process effectively captures critical changes at both the beginning and end stages, leading to more efficient and high-quality image generation.

\section{Ablation Study on Stable Diffusion}
\begin{figure}[t]
  \centering
\begin{subfigure}{\linewidth}
    \includegraphics[width=0.9\linewidth]{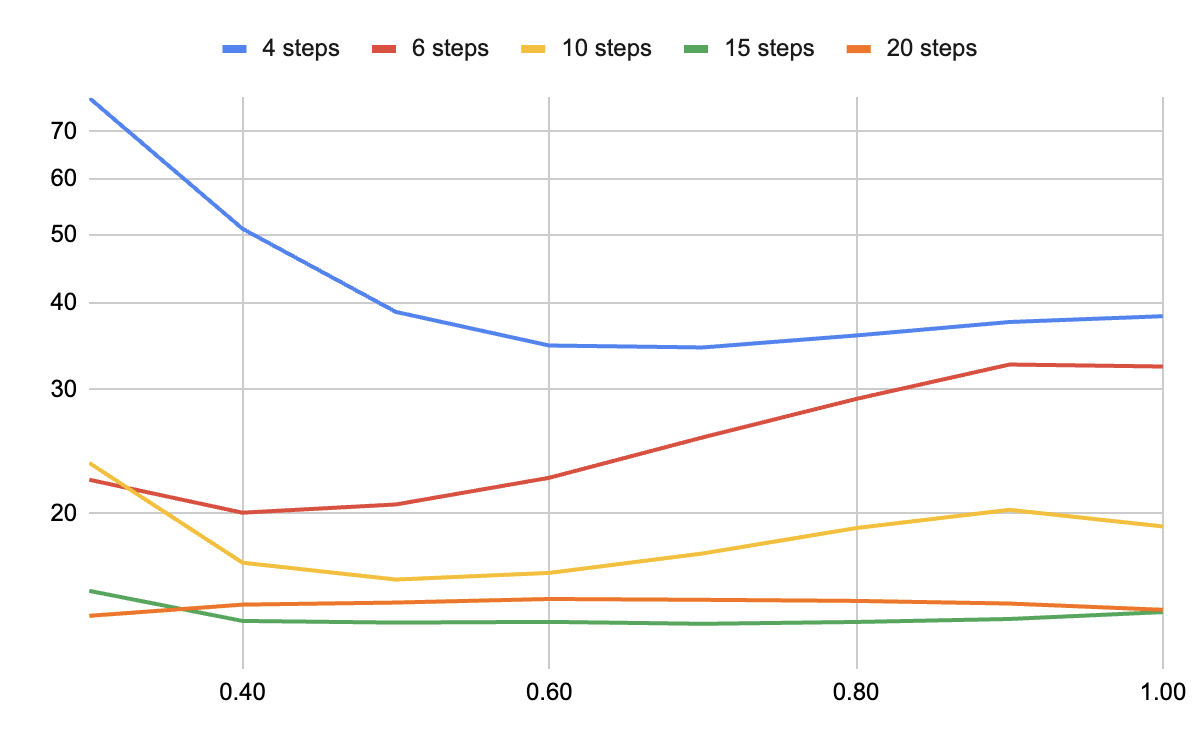}
    \caption{FID}
    \label{fig:short-a}
  \end{subfigure}
  \begin{subfigure}{\linewidth}
    \includegraphics[width=0.9\linewidth]{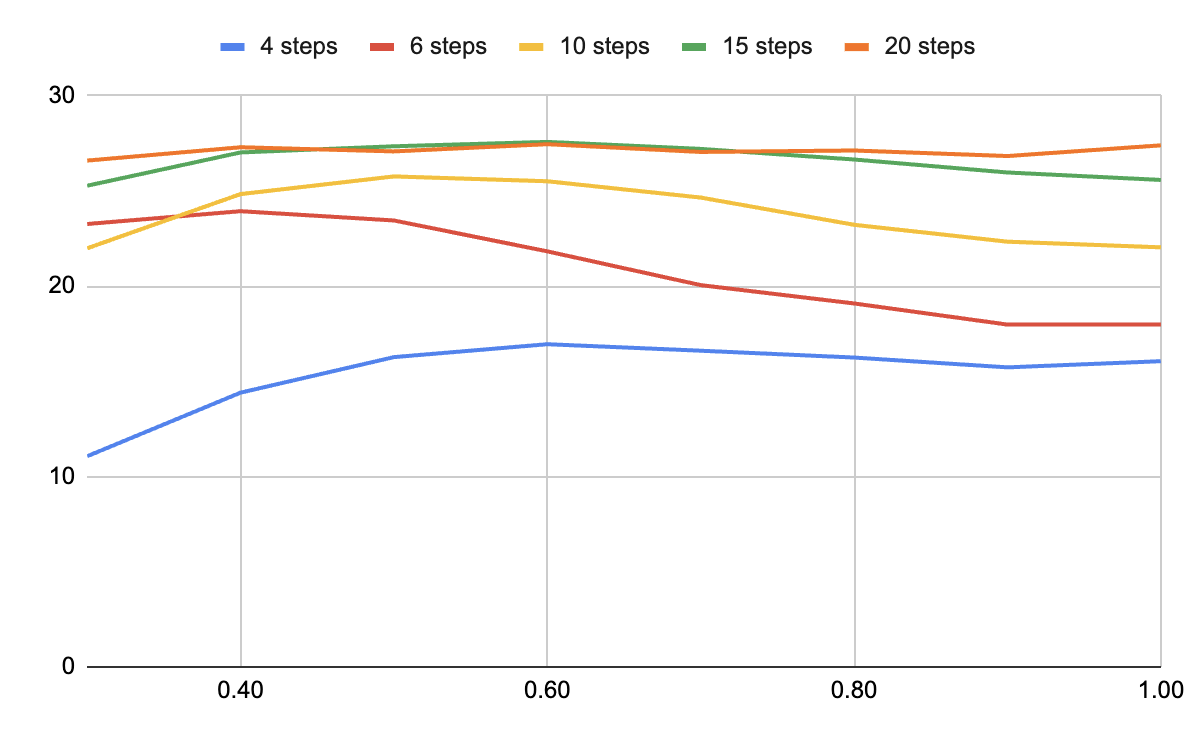}
    \caption{IS}
    \label{fig:short-b}
  \end{subfigure}
  \caption{FID ($\downarrow$) and IS ($\uparrow$) scores for Stable Diffusion with Beta Sampling $Beta(\alpha,\beta)$ of various hyperparameter $\alpha=\beta$ and various time steps.}
  \label{fig:ablation_alpha_sd}
\end{figure}

~\Cref{fig:ablation_alpha_sd} demonstrates the impact of the hyperparameter $\alpha=\beta$ on FID and IS performance in Stable Diffusion, similar to Fig. 7 in the main manuscript. When $\alpha=\beta=1$, the distribution is uniform; as these values decrease, emphasis on the middle stages is reduced. The observed performance shifts in FID and IS as the hyperparameter deviates from the uniform distribution demonstrate the efficacy of Beta Sampling.
The optimal parameter for maximizing FID and IS performance varied with the number of sampled steps. To achieve overall performance enhancement, we set $\alpha = \beta = 0.6$ for Stable Diffusion.

\begin{table}
    \centering
\begin{tabular}{clrr}
\toprule
\multicolumn{1}{c}{Steps} & Sampling Strategies                                                       & FID ($\downarrow$) & IS ($\uparrow$) \\
\midrule
                          & Uniform& \textit{23.19}& \textit{21.59}\\
                          & LogSNR&  37.96& 17.36\\
\multicolumn{1}{c}{4}     & Time quadratic& 29.98 & 19.50\\                 
                          & AutoDiffusion & \textbf{18.53} & \textbf{24.26}\\
                          & \textbf{Ours (Beta)}&  27.34& 19.80\\
                          \midrule
                          & Uniform&17.76 & \textbf{24.73}\\
                          & LogSNR& 17.08 & 23.86\\
\multicolumn{1}{c}{6}     & Time quadratic& 16.25 & 24.19\\                 
                          & AutoDiffusion & \textit{16.15} &\textit{24.65} \\
                          & \textbf{Ours (Beta)}& \textbf{15.83} & 24.53\\
                          \midrule
                          & Uniform&16.20& 26.35\\
                          & LogSNR& 13.98 & \textit{26.88}\\
\multicolumn{1}{c}{10}    & Time quadratic& 13.79 & 26.60\\                 
                          & AutoDiffusion & \textbf{12.61} & 26.69\\
                          & \textbf{Ours (Beta)}& \textit{13.65} & \textbf{27.10}\\
                          \midrule
                          & Uniform&14.39 & 27.35\\
                          & LogSNR& 14.73 & 27.49\\
\multicolumn{1}{c}{20}    & Time quadratic& \textit{13.99} &\textit{27.51} \\                 
                          & AutoDiffusion & \textbf{13.49} & 26.43\\
                          & \textbf{Ours (Beta)}& 14.13 & \textbf{27.67}\\
                          \bottomrule
\end{tabular}
    \caption{FID ($\downarrow$) and IS ($\uparrow$) scores for Stable Diffusion with DPM Solver across various number of time steps and sampling strategies. In this table, Beta Sampling parameters are set to $\alpha=0.5$ and $\beta=0.9$. Bold indicates the \textbf{best} performance values, while italics mark the \textit{second-best}.}
    \label{tab:result_sd_dpm}
\end{table}

\begin{figure}[t]
  \centering
    \includegraphics[width=\linewidth]{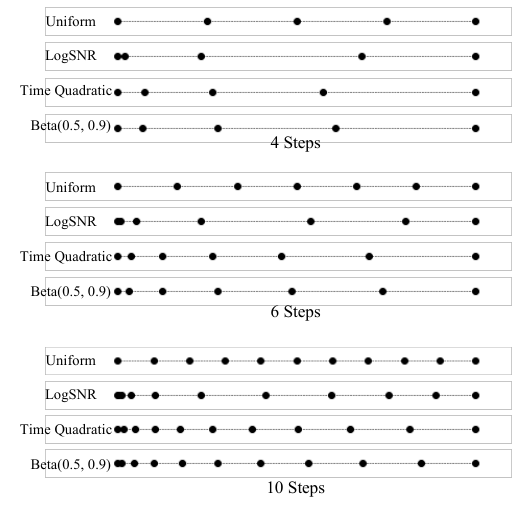}
  
  \caption{Sampled step distribution of Stable Diffusion with DPM solver. Beta Sampling reduces the wide gap in the early stage(right side) while maintaining step density in the later stage(left side).}
  \label{fig:dpm-solver-plot}
\end{figure}

\section{Additional Experiments on Stable Diffusion with DPM Solver}
In \cref{tab:result_sd_dpm}, we present additional experiments on Stable Diffusion using the DPM solver, which supports various skip types. We tested two popular skip types as well as uniform skip types, and found that the Beta Sampling parameters $\alpha=0.5$ and $\beta=0.9$ work effectively for the DPM solver. 
The premise of Beta Sampling is to primarily sample steps at both ends of the process, where significant changes are more likely to occur, rather than the middle portion. The relative emphasis on the initial and latter parts can be adjusted using the weights of $\alpha$ and $\beta$. By applying greater weighting to the latter part, as derived from previous research, the effectiveness of Beta Sampling can be further enhanced.

\end{document}